\RequirePackage{fix-cm}
\ifdefined\pdfminorversion\pdfminorversion=7\fi
\documentclass[11pt]{article}

\usepackage[preprint]{acl}

\usepackage{times}
\usepackage{latexsym}
\usepackage[T1]{fontenc}
\usepackage[utf8]{inputenc}
\usepackage{microtype}
\usepackage{courier}
\usepackage{graphicx}

\usepackage{amsmath,amssymb}
\usepackage{booktabs}
\usepackage{multirow}
\usepackage[table]{xcolor}
\usepackage{enumitem}
\usepackage{float}
\floatstyle{ruled}
\newfloat{algorithm}{tbp}{loa}
\floatname{algorithm}{Algorithm}
\usepackage{tikz}
\usetikzlibrary{positioning,arrows.meta,fit,backgrounds,calc,shapes.misc}
\usepackage{makecell}
\usepackage{tabularx}
\usepackage{adjustbox}
\usepackage{nicefrac}
\usepackage{xspace}

\newcommand{\method}{\textup{\textsc{Relay}}\xspace}

\newcommand{\pit}{\pi_\theta}
\newcommand{\piT}{\pi_{\mathrm{T}}}

\newcommand{\ctx}{\rho}            
\newcommand{\cut}[1][t]{\kappa_{#1}}  
\newcommand{\Dt}{d_t}              

\newcommand{\best}[1]{\textbf{#1}}
\newcommand{\second}[1]{\underline{#1}}
\definecolor{scoregain}{HTML}{1B7837}
\definecolor{scoreloss}{HTML}{B3261E}
\newcommand{\up}[1]{\,{\scriptsize\textcolor{scoregain}{$\uparrow$#1}}}
\newcommand{\dn}[1]{\,{\scriptsize\textcolor{scoreloss}{$\downarrow$#1}}}
\definecolor{oursrow}{RGB}{226,238,252}
\definecolor{relayblue}{HTML}{2a78d6}
\definecolor{grpoorange}{HTML}{eb6834}
\definecolor{aqua}{HTML}{1baf7a}
\definecolor{inkgray}{HTML}{52514e}

\usepackage{tcolorbox}
\definecolor{arxivblue}{HTML}{285779}
\hypersetup{
  pdftitle={RELAY: Selective Self-Distillation with Adaptive Teacher Retirement for Rubric-Based Reinforcement Learning},
  pdfauthor={Kunwang Zheng, Xinyu Sun, Yu Cao, Shuocheng Li, Lei Cheng, Ying Liu, Qilong Zhang, Jinlong Chen, Hua Yang, Qianglong Chen},
  pdfsubject={Selective self-distillation and teacher retirement for rubric-based reinforcement learning},
  citecolor=arxivblue, linkcolor=arxivblue, urlcolor=arxivblue
}

\begin{document}
\onecolumn
\thispagestyle{plain}

\begin{center}
  {\Large\bfseries
  \method: Selective Self-Distillation with Adaptive Teacher Retirement\\
  for Rubric-Based Reinforcement Learning\par}
  \vspace{16pt}
  {\fontsize{11}{14}\selectfont
  Kunwang Zheng\textsuperscript{1,*},
  Xinyu Sun\textsuperscript{2,*},
  Yu Cao\textsuperscript{3},
  Shuocheng Li\textsuperscript{3},
  Lei Cheng\textsuperscript{3}\\[2pt]
  Ying Liu\textsuperscript{3},
  Qilong Zhang\textsuperscript{3},
  Jinlong Chen\textsuperscript{3},
  Hua Yang\textsuperscript{3},
  Qianglong Chen\textsuperscript{3,\dag}\par}
  \vspace{9pt}
  {\fontsize{10}{12}\selectfont
  \textsuperscript{1}University of Science and Technology of China\\
  \textsuperscript{2}Fudan University\qquad
  \textsuperscript{3}Alibaba Group\par}
  \vspace{5pt}
  {\fontsize{9}{11}\selectfont
  \textsuperscript{*}Equal contribution.\qquad
  \textsuperscript{\dag}Corresponding author.\par}
\end{center}

\vspace{16pt}
\begin{center}
\begin{minipage}{0.95\textwidth}
\begin{tcolorbox}[
  colback=white, colframe=arxivblue, boxrule=0.5pt,
  arc=5pt, left=11pt, right=11pt, top=9pt, bottom=9pt,
  before skip=0pt, after skip=0pt
]
  \renewenvironment{abstract}{%
    {\centering\fontsize{11}{13}\selectfont\bfseries Abstract\par}%
    \vspace{5pt}\fontsize{10}{12}\selectfont\noindent\ignorespaces
  }{\par}
  \begin{abstract}
Rubric-based reinforcement learning combines scores for written criteria into a scalar reward, losing information about which criteria a response satisfies.
A teacher that sees the rubric provides token-level supervision.
We study which rollouts should receive this supervision and when to retire the teacher.
We introduce \method, which adds self-distillation from a frozen copy of the initial model to GRPO.
It selects rollouts with positive advantage and those with near-zero advantage and low rubric scores.
When progress in the distillation loss slows, it retires the teacher and continues with GRPO alone.
The stopping rule is calibrated on one model--domain pair and reused unchanged in three others.
\method achieves the highest in-domain score among the compared methods in all four settings, improving on GRPO by 4.62 and 6.02 points on RaR-Medicine for Qwen3-1.7B and Qwen3-4B.
On Qwen3-1.7B/Medicine, removing the teacher from a shared checkpoint yields higher later scores than retaining it, reducing the distillation coefficient, or refreshing it.
In the same setting, distilling selected rollouts that cover about half the response tokens outperforms distilling all rollouts after matching the average distillation coefficient.
\end{abstract}

  \vspace{8pt}
  {\fontsize{9}{11}\selectfont
  \textbf{Email:}
  \href{mailto:kunwang392@gmail.com}{\texttt{kunwang392@gmail.com}}\quad
  \href{mailto:sunxy23@m.fudan.edu.cn}{\texttt{sunxy23@m.fudan.edu.cn}}\par
  \hspace*{30pt}\href{mailto:qianglong.cql@alibaba-inc.com}{\texttt{qianglong.cql@alibaba-inc.com}}\par}
  \vspace{7pt}
  {\centering\fontsize{10}{12}\selectfont\sffamily\bfseries
  \href{https://github.com/kunwang392/RELAY/tree/master}{%
    \makebox[12pt]{\vphantom{Project: github.com/kunwang392/RELAY}%
    \smash{\tikz[baseline=-2pt, x=1pt, y=1pt]{%
      \draw[arxivblue, line width=0.55pt] (0,0) circle (5.5);
      \draw[arxivblue, line width=0.55pt] (0,0) ellipse (2.5 and 5.5);
      \draw[arxivblue, line width=0.55pt] (-5.5,0) -- (5.5,0);
    }}}}\par}
\end{tcolorbox}
\end{minipage}
\end{center}

\vspace{12pt}
\begin{center}
\begin{minipage}{\textwidth}
  \centering
  \captionsetup{hypcap=false}
  \includegraphics[width=0.74\textwidth]{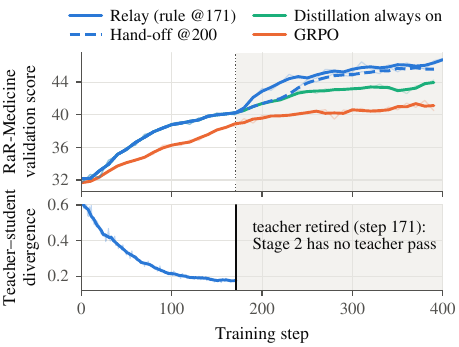}
  \captionof{figure}{Choosing when to retire the teacher on Qwen3-1.7B/Medicine.
  Top: validation scores during training. The three runs with distillation share the same history until their teachers are removed.
  Bottom: mean divergence on the tokens selected for distillation. Its decrease slows before the rule fires at step 171.}
  \label{fig:teaser}
\end{minipage}
\end{center}

\twocolumn

\section{Introduction}
\label{sec:intro}

Rubric-based reinforcement learning trains language models on open-ended tasks by rewarding responses that satisfy instance-specific criteria~\citep{gunjal2025rar,viswanathan2025rlcf,arora2025healthbench}.
An LLM judge combines the criteria into a scalar reward for GRPO~\citep{shao2024deepseekmath}, discarding detail about which facts to include and which errors to avoid.
Self-distillation with privileged context~\citep{zhao2026opsd,hubotter2026sdpo,ye2026opcd} provides token-level guidance: a teacher that reads the rubric scores the student's own rollouts, while the student sees only the question~\citep{bablani2026rupi,rezaei2026rgsd,gu2026rcsd}.
Hybrid methods add this supervision to GRPO.
Our main baseline, CriPO, distills the rollout with the highest advantage in each group toward a teacher based on the current policy, and decays its coefficient to zero late in training~\citep{xia2026cripo}.

This combination leaves two choices: which rollouts the teacher supervises, and how long it stays.
Advantage compares responses within a group, while the rubric reward scores each response against its criteria.
A response can have a low score and zero advantage when its group ties or its reward equals the group mean.
The GRPO reward term then provides no policy gradient.
Prior work on verifiable rewards distills groups in which every rollout fails~\citep{han2026rstg}; graded rubric rewards can tie at any score.
For the initial Qwen3-1.7B policy, 10.9\% of RaR-Medicine groups are tied, a third of them at a nonzero score.

Fixed decay schedules and separate budgets for distillation and RL choose a stopping time before training starts~\citep{xia2026cripo,yang2026rlsd,wang2026papergym,wang2026opdbeforerl}.
On Qwen3-1.7B/Medicine, the teacher--student divergence on distilled tokens falls by about two thirds and then flattens (Figure~\ref{fig:teaser}, bottom), during a similar period in which the validation lead over GRPO stops growing (\S\ref{sec:ana_why}).
We use this loss trend to decide when to retire the teacher.

\method trains in two stages (Figure~\ref{fig:overview}).
Stage~1 combines GRPO with distillation from a frozen copy of the initial model.
It selects rollouts with positive advantage, as well as those with near-zero advantage and low rubric scores.
A detector tracks the relative decrease in the distillation loss and retires the teacher once that decrease remains small.
Stage~2 continues with GRPO alone, without the teacher forward pass.

We train Qwen3-1.7B and Qwen3-4B separately on medicine and science and evaluate the four resulting models on five benchmarks with GPT-5.1.
\method has the highest in-domain score among the compared methods in all four settings.
The detector, calibrated once on Qwen3-1.7B/Medicine, fires between steps 122 and 171 across the four settings with the same parameters.
On Qwen3-1.7B/Medicine, removing the teacher from a shared checkpoint at step~200 yields higher later scores than retaining it, reducing the distillation coefficient, or refreshing it (\S\ref{sec:ana_when}).
Selecting rollouts that cover about half the response tokens also gives shorter responses and higher scores than distilling every rollout after matching the average distillation coefficient (\S\ref{sec:ana_select}).
Our contributions are:
\begin{itemize}
  \item A rule that selects rollouts for distillation using advantages and rubric rewards, including those with low scores and near-zero advantage.
  \item A stopping rule based on the distillation loss, computed during training without additional evaluation.
  \item An evaluation across four model--domain settings, with analyses of rollout selection and teacher retirement on Qwen3-1.7B/Medicine.
\end{itemize}

\section{Related Work}
\label{sec:related}

\paragraph{Rubrics as rewards and context.}
Rubric-based RL scores each criterion and combines the results into one reward~\citep{gunjal2025rar,viswanathan2025rlcf}.
Later work studies how criteria are written and how models can game them~\citep{rezaei2025onlinerubrics,li2026rubrichub,shao2025drtulu,lan2026arlrr,tan2026papo,mahmoud2026rubrichacking}.
Several methods use the criteria to provide more detailed supervision.
RTT adds a reward for each token from a trained relevance discriminator~\citep{xu2026rtt}.
CoRT reweights advantages using the difference in log probabilities between prompts with and without criteria~\citep{zhang2026cort}.
RuscaRL puts the rubric in the policy's context and withdraws it on a fixed schedule~\citep{zhou2025ruscarl}; HeRL feeds unmet criteria back to the policy for revision~\citep{zhang2026herl}.
Both give the policy criteria that are absent at ordinary inference.
\method keeps the student's context, reward, and advantage unchanged. The teacher conveys rubric information through its token probabilities.

\paragraph{Self-distillation with rubrics.}
On-policy distillation trains a student on its own samples using a teacher's token distributions~\citep{agarwal2024gkd,gu2023minillm,li2026rethinkingopd,song2026opdsurvey}.
Recent methods derive the teacher from the student model and give it privileged context~\citep{zhao2026opsd,hubotter2026sdpo,ye2026opcd}.
Several studies report early gains followed by degradation as self-distillation continues~\citep{kim2026sddegrade,zhu2026manyfaces,pan2026rlcsd}.
Three methods give the teacher a rubric: one uses an EMA of the student, or the frozen base when training with LoRA~\citep{bablani2026rupi}; another uses a frozen base model~\citep{rezaei2026rgsd}; a third learns the rubric supplied to the teacher~\citep{gu2026rcsd}.
In all three, teacher supervision replaces the judge or verifier.
We study its use alongside rubric-based RL. In our experiments, distillation alone scores below GRPO on all four tests in the training domains (Table~\ref{tab:main}).

\paragraph{Rollout selection.}
Hybrid methods distill rollouts with positive advantage~\citep{liu2026sdpg}, incorrect rollouts routed to a separate branch~\citep{li2026srpo}, annotated spans~\citep{wang2026steps}, or the rollout with the highest advantage in each group~\citep{xia2026cripo}.
Other methods use a teacher signal to modify token advantages~\citep{yang2026rlsd,kim2026rlrt,pan2026rlcsd}.
Apart from CriPO, these designs assume binary verifiable rewards.
Graded rubric rewards also allow groups to tie at nonzero scores.
Prior work discards tied groups and oversamples to refill the batch~\citep{yu2025dapo,pan2026rlcsd}, adds an advantage with separate normalization that still provides a signal when the outcome advantage is zero~\citep{tan2026papo}, or uses a stronger external teacher to supervise groups that score zero~\citep{han2026rstg}.
\method uses the initial model as its teacher. Its extra branch selects rollouts with near-zero advantage whose rewards fall below a running threshold, covering ties at both zero and nonzero rewards.
RLCSD normalizes its loss separately over tokens with and without advantage modulation~\citep{pan2026rlcsd}; STEPS caps token coverage and decays the distillation weight~\citep{wang2026steps}.
We average over all response tokens and compare our selection rule with distilling all rollouts at a similar average distillation coefficient (\S\ref{sec:ana_select}).

\paragraph{Stopping teacher supervision.}
Prior methods end supervision with a fixed decay schedule~\citep{xia2026cripo,liu2026sdpg,wang2026steps,yang2026rlsd}, keep a positive coefficient throughout training~\citep{han2026rstg}, or set the duration of a distillation phase in advance~\citep{wang2026papergym,wang2026opdbeforerl,li2026sequential}.
OPD-before-RL motivates stopping when the student reaches a plateau below its teacher~\citep{wang2026opdbeforerl}.
Sequential-Beats-Joint tests several switch times and finds benefits from a hard switch after the phase of rapid improvement~\citep{li2026sequential}.
It uses a larger teacher and runs pure distillation before pure RL.
\method combines distillation with RL, then switches to RL alone at a step chosen from the training loss.
RetireOPD is the closest design. In agentic RL, it removes the teacher when the relative change in a sampled-token gap ceases to be negative and the student's validation success rate reaches $0.9$ of the teacher's~\citep{yu2026retireopd}.
Both methods compare windows and remove the teacher while RL continues.
Our rule retires the teacher when the relative decrease in distillation loss stays below a tolerance. It makes this decision from loss values already recorded during training.

\section{Method}
\label{sec:method}

The student generates responses, or \emph{rollouts}, from each prompt without seeing its rubric.
An LLM judge uses the rubric to score these responses.
A frozen copy of the initial model serves as the teacher: it reads the prompt and rubric, then provides token distributions along the student's responses through teacher forcing.
This requires no teacher decoding.
In Stage~1, \method uses rewards and advantages to choose which rollouts receive distillation, while retaining GRPO on all valid response tokens.
We monitor the mean distillation loss on the selected tokens.
When its relative decrease remains small, we remove the teacher and continue with GRPO alone in Stage~2.
Figure~\ref{fig:overview} summarizes the method.
Algorithm~\ref{alg:relay} in Appendix~\ref{app:impl} gives the training loop.

\subsection{Preliminaries}
\label{sec:prelim}

\paragraph{Rubric-based GRPO.}
Each prompt $x$ has a rubric $\ctx=\{(c_j,\omega_j)\}_{j=1}^{M}$ of natural-language criteria with signed weights.
Positive weights specify requirements; negative weights specify penalties.
The policy $\pi_{\theta_{\mathrm{old}}}$ samples a group of $G$ rollouts $y_1,\dots,y_G$, and an LLM judge checks each response against every criterion.
The reward includes both requirements and penalties, using the same rubric as the teacher:
\begin{equation}
  \begin{aligned}
  r_i&=\operatorname{clip}_{[0,1]}\!\left(
    \frac{\sum_{j=1}^{M}\omega_j\,\mathbb{I}[y_i\models c_j]}
         {\sum_{j:\omega_j>0}\omega_j}
  \right),\\
  A_i&=\frac{r_i-\mu_r}{\sigma_r+\epsilon_\sigma},
  \end{aligned}
  \label{eq:reward}
\end{equation}
The indicator is one when a positive requirement is met or a negative penalty is triggered.
Here $\mu_r$ and $\sigma_r$ are the group mean and standard deviation, and $\epsilon_\sigma=10^{-6}$.
Every response token shares $A_i$, and all advantages are zero in a tied group.
Let $\mathcal{V}$ be the positions of valid response tokens in an update batch and $N=|\mathcal{V}|$.
$\mathcal{L}_{\mathrm{GRPO}}$ is the clipped policy objective~\citep{ppo}, averaged over $\mathcal{V}$, plus a KL penalty of weight $\beta$ against the initial model $\pi_0$ on the original context.
We estimate this penalty with the token average of $k_3$ (Appendix~\ref{app:impl}).
Each rollout step supplies $U=2$ gradient updates, each over half the sampled prompts.

\subsection{The teacher and its objective}
\label{sec:teacher}

\paragraph{A frozen teacher.}
The teacher keeps the initial model's weights and receives the prompt with the full signed rubric, $\piT(\cdot\mid x,\ctx,y_{<h})=\pi_0(\cdot\mid[x;\ctx],y_{<h})$.
An appended instruction tells it to satisfy the requirements and avoid the penalties (Appendix~\ref{app:impl}).
Its predictions stay fixed for a given prompt and prefix, while the prefixes sampled by the student change during training.
The initial model also serves as the GRPO KL reference, so the teacher requires no additional model weights.

\paragraph{Distillation objective.}
At token position $h$ in rollout $y_i$, let $K_{i,h}$ contain the teacher's top-$k$ next tokens, with $k=64$.
We renormalize the teacher and student distributions on this set and minimize their conditional forward KL:
\begin{equation}
  \ell_{i,h}=\sum_{v\in K_{i,h}}\bar p_{\mathrm{T}}(v)\log\frac{\bar p_{\mathrm{T}}(v)}{\bar p_\theta(v)},
  \label{eq:token_fkl}
\end{equation}
where $\bar p(v)=p(v)/\sum_{v'\in K_{i,h}}p(v')$ for either distribution at that prefix.
The loss matches the teacher's and student's relative probabilities within $K_{i,h}$.
Appendix~\ref{app:conditional_kl} gives its relation to KL over the full vocabulary.
At rollout step $t$ the combined objective is
\begin{gather}
  \mathcal{L}=\mathcal{L}_{\mathrm{GRPO}}+\alpha_t\,\mathcal{L}_{\mathrm{OPSD}},
  \label{eq:objective}\\
  \mathcal{L}_{\mathrm{OPSD}}=\frac{1}{N}\sum_{(i,h)\in\mathcal{V}}w_i\ell_{i,h},
  \label{eq:opsd_loss}
\end{gather}
with the same token mask and denominator as Eq.~\eqref{eq:grpo}.
The binary weights $w_i$ come from \S\ref{sec:select}.
The coefficient $\alpha_t$ becomes zero when the teacher is removed (\S\ref{sec:switch}).

\begin{figure*}[t]
  \centering
  \includegraphics[width=\textwidth]{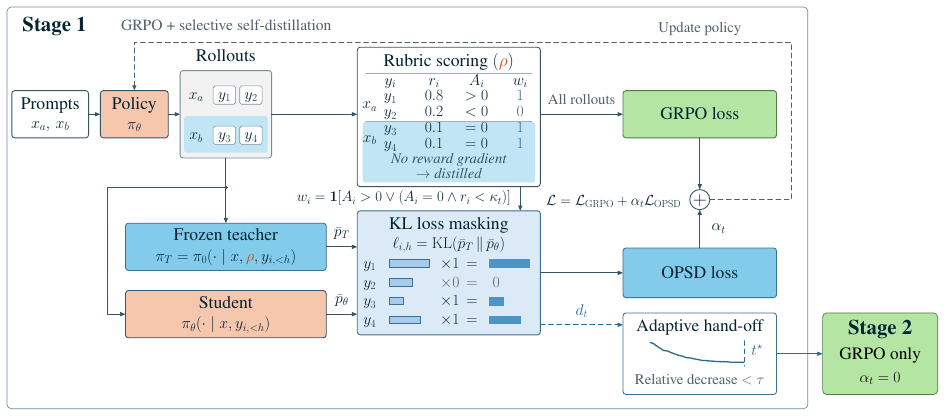}
  \caption{Overview of \method.
  Stage~1 adds distillation to GRPO on selected rollouts.
  When progress in the distillation loss remains small, the rule removes the teacher and Stage~2 continues with GRPO alone.}
  \label{fig:overview}
\end{figure*}

\subsection{Which rollouts to distill}
\label{sec:select}

\paragraph{Using advantages and rewards.}
Advantage compares a response with others to the same prompt; its reward records how well it satisfies the rubric.
A response can score poorly yet have zero advantage, either because all responses tie or because its reward equals the group mean.
We distill rollouts with positive advantage and those with near-zero advantage whose rewards fall below a running cutoff.
The latter receive little or no correction from the GRPO reward term.
The cutoff directs this extra supervision to lower rewards in the current training distribution.
Table~\ref{tab:rollout_types} in Appendix~\ref{app:impl} gives the treatment of each group.

During Stage~1, selected tokens receive distillation weight $\alpha$ and the remaining tokens receive zero weight.
If selection covers a fraction $f$ of all valid response tokens, the average distillation coefficient is $\alpha f$.
Equivalently, the distillation contribution to Eq.~\eqref{eq:objective} is $\alpha f$ times the mean loss on the selected tokens.

\paragraph{Rule.}
A quality mask $q_i$ screens out empty rollouts, those past the length limit, and those with nonfinite rewards or advantages.
The cutoff $\cut$ is an exponential moving average of the 30th percentile of the rewards in the current batch,
$\cut=(1-\lambda)\cut[t-1]+\lambda Q_{0.3}(\{r_i:q_i=1\}_t)$ with $\lambda=0.1$, initialized at the first batch's 30th percentile.
We then select
\begin{equation}
  w_i=q_i\,\mathbb{I}\big[A_i>\epsilon_A\ \lor\ (|A_i|\le\epsilon_A\ \land\ r_i<\cut)\big],
  \label{eq:select}
\end{equation}
with $\epsilon_A=10^{-6}$ as the numerical tolerance for zero advantage.
The percentile cutoff follows shifts in the reward distribution.
Across the four settings, $f$ ranges from $0.37$ to $0.60$, with a mean near $0.49$ in each (Appendix~\ref{app:plateau}).

\subsection{When to retire the teacher}
\label{sec:switch}

\paragraph{Tracking progress.}
We retire the teacher when the relative decrease in the conditional KL loss stays below a tolerance for consecutive window comparisons.

\paragraph{Statistic.}
Let $D_{t,u}$ be the positions chosen for distillation in update $u$ of rollout step $t$.
We number steps from zero and compute each mean using the policy that enters that update.
We record the loss on the selected tokens,
\begin{equation}
  \Dt=\frac{1}{U}\sum_{u=1}^{U}\frac{1}{|D_{t,u}|}\sum_{(i,h)\in D_{t,u}}\ell_{i,h},
  \label{eq:plateau}
\end{equation}
which takes the mean loss on selected tokens within each update, then averages these means equally across updates.
This removes the direct scaling by the fraction of tokens selected.
An update with no selected tokens contributes zero.

\paragraph{Detector.}
Write $g_0,\ldots,g_j$ for the recorded values of $\Dt$, omitting steps without distilled tokens.
Once $2W$ values are available, compare consecutive windows of $W$:
\begin{equation}
  m_j=\frac{1}{W}\sum_{n=j-W+1}^{j}g_n,\qquad
  \delta_j=\frac{m_{j-W}-m_j}{m_{j-W}}.
\end{equation}
The rule fires at the first rollout step $t^\star\ge t_{\min}-1$ at which the last $P$ available values of $\delta_j$ all fall below $\tau$; if that never happens, $t^\star=+\infty$.
A deadline $T_{\max}$ bounds the duration of Stage~1, giving $t_s=\min(t^\star,T_{\max})$.
We use $\alpha_t=\alpha$ through step $t_s$ and set it to zero from step $t_s+1$.
Stage~2 then continues with GRPO and the original KL reference, without the teacher forward pass.

\paragraph{Calibration.}
We use $W=20$, $\tau=0.05$, $P=5$, $t_{\min}=60$ and $T_{\max}=199$.
We chose the window and tolerance by replaying the Qwen3-1.7B/Medicine Stage-1 trace over a grid of values.
All settings were then reused unchanged for the other three model--domain pairs (Appendix~\ref{app:plateau}).

\section{Experiments}
\label{sec:exp}

\begin{table*}[t]
  \centering
  
  \small
  \setlength{\tabcolsep}{5.8pt}
  \begin{tabular}{llllclll}
    \toprule
    & \multicolumn{3}{c}{\textbf{Medicine}} & & \multicolumn{2}{c}{\textbf{Science}} & \\
    \cmidrule(lr){2-4}\cmidrule(lr){6-7}
    \textbf{Method} & RaR-Med. & HealthBench & LLMEval-Med & & RaR-Sci. & ResearchQA & \textbf{Aggregate} \\
    \midrule
    \multicolumn{8}{l}{\emph{Qwen3-1.7B}} \\
    Base model         & 34.87 & 21.91 & 54.11 & & 47.79 & 43.54 & 40.44 \\
    \quad + OPSD       & 40.20\up{5.33} & 18.79\dn{3.12} & 55.71\up{1.60} & & 49.25\up{1.46} & 45.31\up{1.77} & 41.85\up{1.41} \\
    \quad + GRPO       & 42.15\up{7.28} & 23.87\up{1.96} & \second{56.28}\up{2.17} & & 52.79\up{5.00} & 51.11\up{7.57} & 45.24\up{4.80} \\
    \quad + CriPO      & \second{42.37}\up{7.50} & \second{27.97}\up{6.06} & 54.99\up{0.88} & & \second{55.79}\up{8.00} & \best{60.58}\up{17.04} & \second{48.34}\up{7.90} \\
    \rowcolor{oursrow}
    \quad + \method    & \best{46.77}\up{11.90} & \best{29.31}\up{7.40} & \best{57.51}\up{3.40} & & \best{57.00}\up{9.21} & \second{59.73}\up{16.19} & \best{50.06}\up{9.62} \\
    \midrule
    \multicolumn{8}{l}{\emph{Qwen3-4B}} \\
    Base model         & 48.56 & 33.82 & 63.18 & & 62.60 & 51.85 & 52.00 \\
    \quad + OPSD       & 56.13\up{7.57} & 29.17\dn{4.65} & 66.21\up{3.03} & & 63.29\up{0.69} & 52.89\up{1.04} & 53.54\up{1.54} \\
    \quad + GRPO       & 58.41\up{9.85} & 39.53\up{5.71} & \second{68.01}\up{4.83} & & \second{68.36}\up{5.76} & 63.19\up{11.34} & 59.50\up{7.50} \\
    \quad + CriPO      & \second{59.67}\up{11.11} & \second{41.93}\up{8.11} & 67.77\up{4.59} & & 66.17\up{3.57} & \second{63.95}\up{12.10} & \second{59.90}\up{7.90} \\
    \rowcolor{oursrow}
    \quad + \method    & \best{64.43}\up{15.87} & \best{42.63}\up{8.81} & \best{69.14}\up{5.96} & & \best{69.80}\up{7.20} & \best{65.79}\up{13.94} & \best{62.36}\up{10.36} \\
    \bottomrule
  \end{tabular}
  \caption{GPT-5.1 scores averaged over three evaluations with fresh responses (standard deviations: Table~\ref{tab:repeated_eval}).
  Models are trained separately by domain; Aggregate averages the five benchmarks equally.
  Best/second best per model size: \textbf{bold}/\underline{underlined}.
  Arrows show point changes from the same-size base model (\textcolor{scoregain}{$\uparrow$}/\textcolor{scoreloss}{$\downarrow$}: increase/decrease).}
  \label{tab:main}
\end{table*}

\subsection{Setup}
\label{sec:setup}

\paragraph{Data and models.}
We train Qwen3-1.7B and Qwen3-4B~\citep{qwen3} in non-thinking mode on RaR-Medicine and RaR-Science~\citep{gunjal2025rar}, with a separate model for each domain and size.
We evaluate on the corresponding RaR test split and on additional benchmarks from the same domain: HealthBench~\citep{arora2025healthbench} and LLMEval-Med~\citep{zhang-etal-2025-llmeval} for medicine, and ResearchQA~\citep{yifei2025researchqa} for science (Appendix~\ref{app:data}).

\paragraph{Training.}
All RL methods use 64 prompts with 8 responses per rollout step and a nominal budget of 400 steps.
Qwen3.8-27B~\citep{qwen38} provides training rewards and validation scores; Appendix~\ref{app:impl} gives the remaining training settings.
Checkpoint selection and validation curves use the official RaR validation partitions; final GPT-5.1 evaluation uses the separate test partitions.

\paragraph{Evaluation.}
GPT-5.1~\citep{openai2025gpt51}, with reasoning disabled, evaluates responses from the checkpoints in Table~\ref{tab:main} independently of the training judge.
Responses are sampled at temperature 0.7 with the same 5{,}120-token limit as training.
RaR uses the full signed rubric and clips each question's normalized score to $[0,1]$; the three additional benchmarks use adapted protocols (Appendix~\ref{app:eval}).
We evaluate checkpoints near the peaks of validation during training.
For the checkpoints in Table~\ref{tab:main}, we run three evaluations with newly sampled responses and report the mean score.

\paragraph{Baselines.}
We compare with the base model, standalone OPSD~\citep{zhao2026opsd} from the same frozen teacher, GRPO~\citep{shao2024deepseekmath}, and our reproduction of CriPO~\citep{xia2026cripo}.
CriPO distills the rollout with the highest advantage in each group from a teacher based on the current policy, corrects token advantages, and decays the distillation coefficient late in training.
We use $\beta=0.01$ for CriPO and compare it with the published value of $\beta=0.05$ in Appendix~\ref{app:cripo}.

\begin{figure*}[t]
  \centering
  \includegraphics[width=\textwidth]{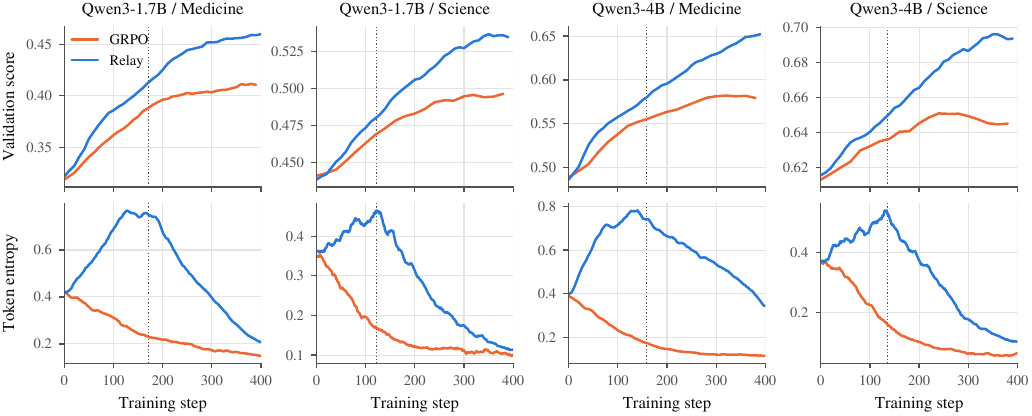}
  \caption{Validation scores on the 0--1 scale (top) and rollout token entropy (bottom) for GRPO and \method across the four settings. Curves show 10-step averages, with the starting validation scores aligned. Dotted lines mark the step at which \method removes the teacher.}
  \label{fig:dynamics}
\end{figure*}

\subsection{Main results}
\label{sec:main}

\paragraph{Results on the RaR test splits.}
\method has the highest in-domain score among the evaluated methods (Table~\ref{tab:main}).
The gains over GRPO are 4.62 and 6.02 points on RaR-Medicine, and 4.21 and 1.44 points on RaR-Science, for the 1.7B and 4B models, respectively.
\method also outperforms the CriPO reproduction in all four in-domain settings and has the highest average across the five benchmarks at both sizes.
In the first evaluation, the 95\% confidence intervals from bootstrapping questions exclude zero for all eight paired differences on the RaR tests (Figure~\ref{fig:gains} in Appendix~\ref{app:eval}).

\paragraph{Additional benchmarks.}
Across all five benchmarks, \method has the highest score on four at 1.7B and on all five at 4B.
It improves HealthBench over GRPO by 5.4 and 3.1 points.
On LLMEval-Med it scores highest at both sizes, 57.51 and 69.14.
Appendix~\ref{app:case} follows one science question through all three methods.

\paragraph{Standalone distillation.}
OPSD improves on the base model in all four in-domain settings but trails GRPO, and its validation scores peak at steps 20--185 before declining (Appendix~\ref{app:baselines}).
On HealthBench, OPSD falls below the base model at both sizes, while \method improves over both the base model and GRPO.
These outcomes favor combining the two forms of supervision: the teacher guides token probabilities, and the rubric reward favors complete responses with higher scores.

\section{Analysis}
\label{sec:analysis}

We examine teacher removal, its timing, the loss plateau, and rollout selection.
The controlled comparisons use Qwen3-1.7B/Medicine; we also report training curves and stopping times across all four settings.
Scores quoted below use the 0--100 scale and come from the training judge unless GPT-5.1 is named.

Across the four settings, \method maintains higher entropy than GRPO during most of Stage~1 (Figure~\ref{fig:dynamics}).
After teacher removal, entropy declines while validation scores continue to improve.

\subsection{Does retiring the teacher help?}
\label{sec:ana_when}

\paragraph{Branches from one checkpoint.}
For this comparison, we keep distillation active through step~200 and branch from one checkpoint into four runs.
They share the preceding training history and differ only in how they use the teacher: removed, kept at $\alpha=0.01$, kept at $\alpha=0.003$, or replaced once by a copy of the current policy at the original coefficient (Table~\ref{tab:branches}).
Gains are measured against the common validation baseline of 40.62, averaged over steps 160--200.
Over steps 210--260 the four branches lie within 0.8 points of each other, and the reduced coefficient is briefly ahead.
In each reported interval from steps 270--300 onward, removing the teacher gives the largest average gain among the four branches.
Over steps 360--390, removal gives an average gain of 5.06 points, compared with 2.88 for keeping the teacher at full strength (Table~\ref{tab:branches}).
Reducing the coefficient weakens teacher supervision; refreshing the teacher changes its weights while retaining the distillation objective.
Both alternatives improve later scores over keeping the original teacher at full strength.
Removing the teacher improves them further, favoring a transition to GRPO alone after the shared training phase.

\begin{table}[!t]
  \centering
  \footnotesize
  \setlength{\tabcolsep}{3pt}
  \begin{tabular}{@{}lcccc@{}}
    \toprule
    Branch at step 200 & 210--260 & 270--300 & 310--350 & 360--390 \\
    \midrule
    \rowcolor{oursrow}
    Teacher removed            & 2.31 & \textbf{4.14} & \textbf{4.71} & \textbf{5.06} \\
    Kept at $\alpha{=}0.003$    & \textbf{2.49} & 3.45 & 3.73 & 4.11 \\
    Teacher refreshed          & 1.77 & 3.22 & 3.55 & 3.76 \\
    Kept at $\alpha{=}0.01$     & 1.70 & 2.49 & 2.67 & 2.88 \\
    \bottomrule
  \end{tabular}
  \caption{Validation gains for four runs that branch at step~200, evaluated by the training judge and averaged over the indicated steps. All use the mean score over steps 160--200 as the reference. Bold marks the column maximum.}
  \label{tab:branches}
\end{table}

\paragraph{Entropy after teacher removal.}
Entropy declines after teacher removal at steps~140, 171 and~200 (Table~\ref{tab:handoff_entropy} in Appendix~\ref{app:branches}).
By steps 281--300, entropy is 0.32--0.42 after teacher removal, compared with 0.60 under continued distillation.
GRPO has lower entropy throughout these intervals.
After retirement, the distribution over next tokens becomes more concentrated as validation scores continue to improve (Figure~\ref{fig:dynamics}).
The CriPO reproduction also maintains higher entropy later in training (Figure~\ref{fig:cripo_beta} in Appendix~\ref{app:cripo}).

\subsection{Stopping time across settings}
\label{sec:ana_trigger}

\paragraph{Fixed and adaptive stopping.}
Table~\ref{tab:schedule} compares the stopping rule with removal at steps~140 and~200, continued distillation, and GRPO.
All three removal times improve the peak validation gain over continued distillation.
The rule fires at step~171 and reaches a peak gain of 14.8 points.
Its GPT-5.1 score is 46.77, compared with 45.81 and 46.38 for removal at steps~140 and~200, respectively, and 42.15 for GRPO.
The three removal times have the same ranking under the training judge and GPT-5.1.

\begin{table}[!ht]
  \centering
  \footnotesize
  \setlength{\tabcolsep}{4pt}
  \begin{tabular}{@{}lccc@{}}
    \toprule
    Schedule & Best $\Delta$ & Peak step & GPT-5.1 \\
    \midrule
    GRPO                         & 9.8  & 370 & 42.15 \\
    Always on                    & 12.3 & 390 & 44.17 \\
    \rowcolor{oursrow}
    Hand-off @140                & 13.3 & 350 & 45.81 \\
    \rowcolor{oursrow}
    Hand-off @200                & 14.1 & 370 & 46.38 \\
    \rowcolor{oursrow}
    \method{} (rule @171)        & \textbf{14.8} & 400 & \textbf{46.77} \\
    \bottomrule
  \end{tabular}
  \caption{Teacher removal schedules on Qwen3-1.7B/Medicine.
  Best $\Delta$: largest validation gain over step~0, scored by the training judge and smoothed over three steps; Peak step: its location.
  GPT-5.1: RaR-Medicine test score, with GRPO and \method taken from Table~\ref{tab:main}.
  Always on keeps $\alpha=0.01$ throughout.}
  \label{tab:schedule}
\end{table}

\begin{figure*}[t]
  \centering
  \includegraphics[width=\textwidth]{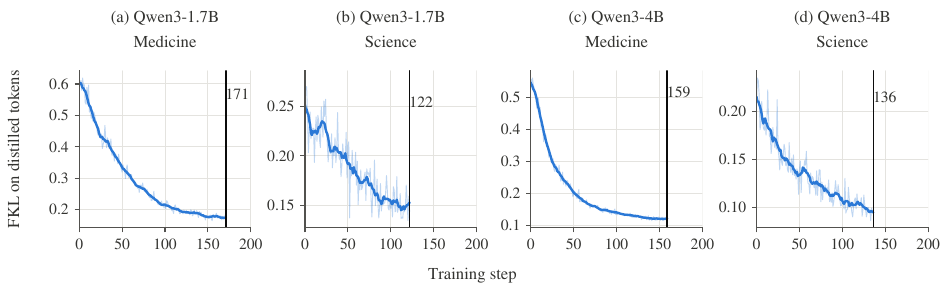}
  \caption{Mean conditional forward KL on tokens selected for distillation in the four model--domain settings. Vertical lines mark the steps at which the stopping rule fires.}
  \label{fig:fkl_traces}
\end{figure*}

\paragraph{Transfer.}
We calibrate the detector on Qwen3-1.7B/Medicine and reuse its settings for the other three model--domain pairs.
It fires at steps~171 and~122 for Qwen3-1.7B on Medicine and Science, and at steps~159 and~136 for Qwen3-4B on Medicine and Science (Figure~\ref{fig:fkl_traces}).
All four triggers precede the deadline $T_{\max}=199$.
The absolute divergence levels differ by more than a factor of two across these settings.
The detector divides the decrease between consecutive window means by the earlier mean, so the same 5\% tolerance measures progress relative to each trace's loss level.
A constant rescaling of the loss cancels in this ratio.
Appendix~\ref{app:plateau} reports sensitivity to the window length and tolerance.

\subsection{What the plateau measures}
\label{sec:ana_why}

\paragraph{The signal and the gain flatten together.}
The monitored divergence and the lead over GRPO flatten over a similar period (Figure~\ref{fig:teaser}).
In the trace with distillation continued through step~200, the conditional KL between teacher and student falls from 0.55 at the start of Stage~1 to 0.18 by step~150 and holds there to step~200.
The lead over GRPO stops growing in the same stretch: it reaches 2.9 points by step~110 and goes no higher through step~200.
The two traces connect the loss plateau to the end of the period in which distillation expands the validation lead.

\paragraph{A remaining gap at the plateau.}
When generating with the rubric, the frozen teacher scores 90.5, compared with 40.4 for the student at step~140.
Meanwhile, top-64 overlap changes little, from 0.40 to 0.44 over Stage~1, despite the reduction in conditional KL.
By the end of Stage~1, their top-1 predictions still differ at 56\% of positions (Appendix~\ref{app:plateau}).
Because the loss is normalized within the teacher's top-64 set, it measures agreement in the relative probabilities of those tokens (Appendix~\ref{app:conditional_kl}).
It can therefore decrease substantially while the overlap between the models' preferred tokens changes little.
The teacher remains strong when it sees the rubric, while progress through the current distillation objective has slowed.

\paragraph{Why retire the teacher at the plateau?}
At the plateau, distillation still asks the student to match the teacher's token preferences even though the validation lead has stopped growing.
On selected rollouts, the cosine between the distillation and GRPO gradients ranges from $-0.11$ to $+0.05$ at checkpoints from step~100 onward (Table~\ref{tab:grad} in Appendix~\ref{app:grad}).
Continued teacher matching therefore supplies a direction with little consistent alignment to GRPO.
Together with the loss plateau, this suggests limited benefit from retaining the additional objective and motivates letting GRPO guide the remaining training.
The shared-checkpoint comparison in \S\ref{sec:ana_when} supports this interpretation: removing the teacher gives higher later scores than keeping it, reducing its coefficient, or refreshing it.

\subsection{Which rollouts to distill}
\label{sec:ana_select}

\begin{table}[!t]
  \centering
  \footnotesize
  \begin{tabular}{@{}lcccc@{}}
    \toprule
    Distilled & Tok. & Best $\Delta$ & Len. & Tr. \\
    \midrule
    \rowcolor{oursrow}
    Pos.\,+\,low zero & .49 & \textbf{7.9} & 1{,}482 & 0.04 \\
    Pos.\,+\,all zero  & .54 & 7.6 & 1{,}643 & 0.13 \\
    Positive only        & .42 & 7.5 & 1{,}309 & 0.02 \\
    Pos.\ + negative     & .91 & 7.3 & 1{,}752 & 0.35 \\
    All rollouts         & .92 & 6.8 & 2{,}262 & 2.39 \\
    All, $\alpha{=}.005$ & .98 & 7.1 & 1{,}978 & 1.17 \\
    \midrule
    GRPO                 & --- & 5.6 & 1{,}023 & 0.02 \\
    \bottomrule
  \end{tabular}
  \caption{Rollout selection in Stage~1 (Qwen3-1.7B/Medicine).
  Tok.: distilled share of response tokens; low zero: rollouts with $|A_i|\le\epsilon_A$ and $r_i<\cut$.
  Best $\Delta$: peak validation gain during Stage~1, scored by the training judge.
  Len., Tr.: rollout length and truncation (\%) at steps 151--200.
  Distilling all rollouts with $\alpha=0.005$ matches \method's average distillation coefficient.}
  \label{tab:selection}
\end{table}

Distillation trains the student to match the teacher's relative token probabilities at the sampled prefixes.
Differences between valid continuations can also contribute to this loss.
We use the reward and advantage to decide where to apply this additional supervision.

\paragraph{Positive and negative advantage.}
The GRPO reward term favors responses with positive advantage.
Such a response can still miss requirements in the rubric, so the teacher can refine the student's predictions using the full rubric.
For rollouts with negative advantage, the reward term already supplies a signal to reduce their probability.
Teacher forcing conditions on the sampled prefixes.
If a prefix contains an error, later teacher targets describe how to continue or repair that response.
Learning these continuations can encourage elaboration of a response that the reward term is discouraging.
Our rule therefore leaves these rollouts to GRPO.
A rollout can have near-zero advantage simply because its score is close to the group mean, even when rewards within the group differ.
If the score is also low, teacher guidance supplies supervision where the GRPO reward term offers little direction.
The cutoff prioritizes these responses and limits additional matching that could alter behavior already receiving higher rewards.

\paragraph{Rollout coverage and peak gain.}
The comparisons use $\alpha=0.01$, chosen once by an ablation over $\{0.1,0.01,0.001\}$.
Across five checkpoints, the aggregate distillation gradients from rollouts with positive and negative advantage have cosines of 0.97--0.99 (Table~\ref{tab:grad} in Appendix~\ref{app:grad}).
Adding the negative group therefore extends a similar direction of teacher supervision to more trajectories.
In the ablation, this addition increases mean response length from 1{,}309 to 1{,}752 tokens and lowers peak gain from 7.5 to 7.3 points.
Distilling all rollouts gives a peak gain of 6.8 points.

Adding rollouts with low scores and near-zero advantage to those with positive advantage raises peak gain from 7.5 to 7.9 points, while the fraction of selected tokens increases from 0.42 to 0.49.
Extending this branch to all rollouts with zero advantage gives a lower gain of 7.6, increases the selected fraction to 0.54, and increases mean response length from 1{,}482 to 1{,}643 tokens (Table~\ref{tab:selection}).
Across these comparisons, length and truncation rise monotonically with the fraction $f$ of selected tokens (see Figure~\ref{fig:selection} for length curves).

\paragraph{Matching the average coefficient.}
The distillation loss is averaged over all valid response tokens, so selecting a fraction $f$ gives an average coefficient of $\alpha f$.
At $\alpha=0.01$, this coefficient is $0.0092$ for distilling all rollouts and $0.0049$ for \method.
To examine the allocation of supervision at a similar average coefficient, we reduce $\alpha$ for distilling all rollouts to $0.005$.
With a selected fraction of $0.98$, its average coefficient is then about $0.0049$.
\method achieves a peak gain of 7.9 points, compared with 7.1 for this alternative, and produces shorter responses: 1{,}482 versus 1{,}978 tokens (Table~\ref{tab:selection}).
At step~200, \method also scores 1.49 points higher on RaR-Medicine under GPT-5.1.
The advantage of selection therefore persists when the two rules have a similar average distillation coefficient.

\section{Conclusion}
\label{sec:conclusion}

\method selects rollouts for privileged self-distillation using their advantages and rubric rewards, then retires the frozen teacher when progress in the distillation loss slows.
It has the highest in-domain score among the compared methods in all four model--domain settings, with a stopping rule calibrated once and reused unchanged.

On Qwen3-1.7B/Medicine, removing the teacher from a shared checkpoint at step~200 yields higher later scores than retaining it, reducing the distillation coefficient, or refreshing it.
Selecting rollouts that cover about half the response tokens also outperforms distilling every rollout after matching the average distillation coefficient.

The rubric-conditioned teacher still scores substantially higher in the generation diagnostic (\S\ref{sec:ana_why}).
This remaining gap motivates studying objectives that can make further use of the teacher.

\section*{Limitations}
\label{sec:limitations}

\paragraph{Single calibration setting.}
The plateau rule is calibrated on one model--domain pair and reused in three others.
Some settings in the sensitivity analysis do not trigger within the available training traces, so substantially different training dynamics or model scales may require new calibration.

\paragraph{Frozen teacher.}
The stopping rule has been evaluated with a fixed teacher.
Its behavior when the teacher is updated during training remains to be studied.

\paragraph{Binary selection.}
The selection rule gives each rollout a binary weight.
Continuous weights could allocate supervision more finely; their interaction with the stopping rule remains to be studied.

\paragraph{Seeds.}
We train one seed per configuration, so variation across training seeds is unmeasured.
The standard deviations reflect repeated evaluations at fixed checkpoints; the confidence intervals resample questions from the first evaluation.

\section*{Ethical Considerations}
The models in this paper answer medical questions.
Their outputs have not been validated by clinicians and are unsuitable for clinical decision-making.
All datasets are public research benchmarks; we did not collect new data from people.

\bibliography{references}

\clearpage
\appendix
\makeatletter
\setlength{\@fptop}{0pt plus 1fil}
\setlength{\@fpsep}{16pt plus 4pt minus 4pt}
\setlength{\@fpbot}{0pt plus 1fil}
\setlength{\@dblfptop}{0pt plus 1fil}
\setlength{\@dblfpsep}{16pt plus 4pt minus 4pt}
\setlength{\@dblfpbot}{0pt plus 1fil}
\makeatother
\setlength{\textfloatsep}{10pt plus 2pt minus 2pt}
\setlength{\dbltextfloatsep}{10pt plus 2pt minus 2pt}
\setlength{\intextsep}{8pt plus 2pt minus 2pt}
\setlength{\floatsep}{8pt plus 2pt minus 2pt}
\setlength{\dblfloatsep}{8pt plus 2pt minus 2pt}
\captionsetup{font=small}

\section{Implementation Details}
\label{app:impl}

Writing $\eta_{i,h}=\pit(y_{i,h}\mid x,y_{i,<h})/\pi_{\theta_{\mathrm{old}}}(y_{i,h}\mid x,y_{i,<h})$ and $\tilde\eta_{i,h}=\mathrm{clip}(\eta_{i,h},1-\epsilon_{\mathrm{c}},1+\epsilon_{\mathrm{c}})$, the policy term of \S\ref{sec:prelim} is
\begin{equation}
  \begin{aligned}
  \mathcal{L}_{\mathrm{GRPO}}&=-\frac{1}{N}\sum_{(i,h)\in\mathcal{V}}\min(\eta_{i,h}A_i,\tilde\eta_{i,h}A_i)\\
  &\quad+\beta\mathcal{L}_{\mathrm{KL}}.
  \end{aligned}
  \label{eq:grpo}
\end{equation}
Table~\ref{tab:hparams} lists the shared training configuration.
All runs use ROLL~\citep{wang2025roll} with Megatron-Core training and vLLM generation on 8 H200 GPUs (tensor parallel 4, data parallel 2).
During training, we validate on the official RaR validation partitions with the training judge. GPT-5.1 evaluates the separate test partitions.

Algorithm~\ref{alg:relay} gives the training loop, and Table~\ref{tab:rollout_types} summarizes how each class of rollout is treated.

\begin{algorithm}[!ht]
\small
\caption{\method training loop}
\label{alg:relay}
\noindent\textbf{Require:} Initial policy $\pi_0$, total rollout steps $T$\par
\noindent\textbf{Require:} $\alpha,W,\tau,P,t_{\min},T_{\max}$\par
\newcommand{\algline}[2]{\item\hspace*{#1em}\parbox[t]{\dimexpr\linewidth-#1em\relax}{#2}}
\begin{enumerate}[label=\arabic*:,leftmargin=1.8em,labelsep=0.5em,itemsep=0pt,parsep=0pt,topsep=2pt]
\item $\pit\gets\pi_0$; $t_s\gets T_{\max}$; history $H\gets[\,]$
\item \textbf{for} $t=0,1,\ldots,T-1$ \textbf{do}
\algline{1}{Sample groups; judge responses; compute $r_i,A_i$}
\algline{1}{\textbf{if} $t\le t_s$ \textbf{then}}
\algline{2}{Teacher-force $\piT$ on the sampled responses}
\algline{2}{Form quality masks $q_i$}
\algline{2}{Initialize/update $\cut$ if any $q_i=1$}
\algline{2}{Select $w_i$ using Eq.~\eqref{eq:select}}
\algline{2}{Perform two updates with Eq.~\eqref{eq:objective}}
\algline{2}{\textbf{if} this step contains distilled tokens \textbf{then}}
\algline{3}{Append $d_t$ to $H$ (Eq.~\eqref{eq:plateau})}
\algline{3}{\textbf{if} $t+1\ge t_{\min}$ and plateau$(H;W,\tau,P)$ \textbf{then}}
\algline{4}{$t_s\gets\min(t,T_{\max})$}
\algline{1}{\textbf{else}}
\algline{2}{Perform two GRPO updates;\\skip teacher and selection}
\end{enumerate}
\end{algorithm}

\begin{table}[!ht]
  \centering
  \footnotesize
  \setlength{\tabcolsep}{4pt}
  \begin{tabular}{@{}llc@{}}
    \toprule
    Rollout & GRPO reward gradient & Distilled \\
    \midrule
    $A_i>\epsilon_A$                         & towards the rollout & yes \\
    $A_i<-\epsilon_A$                        & away from it        & no \\
    $|A_i|\le\epsilon_A$, $r_i<\cut$          & approximately zero  & yes \\
    $|A_i|\le\epsilon_A$, $r_i\ge\cut$        & approximately zero  & no \\
    \bottomrule
  \end{tabular}
  \caption{GRPO and distillation for each class of rollout.}
  \label{tab:rollout_types}
\end{table}

\begin{table*}[!t]
  \centering
  \footnotesize
  \setlength{\tabcolsep}{4pt}
    \begin{tabular}{@{}p{0.29\textwidth}p{0.66\textwidth}@{}}
    \toprule
    Setting & Value \\
    \midrule
    Rollout budget & 400 steps \\
    Prompts $\times$ responses per step & 64 $\times$ 8 \\
    Policy updates per step & 2 (32 prompts each) \\
    Sampling (train) & $T{=}1$, top-$p{=}1$, top-$k{=}{-}1$ \\
    Prompt / response limit & 1{,}024 / 5{,}120 tokens \\
    Optimizer & Adam, LR $10^{-6}$ constant, no warmup \\
    PPO clip & 0.2 \\
    KL to initial model & $\beta=10^{-3}$, $k_3$ estimator \\
    Training reward & Eq.~\eqref{eq:reward}, judge temperature 0.1 \\
    Advantage & $(r-\mu)/(\sigma+\epsilon_\sigma)$, $\epsilon_\sigma{=}10^{-6}$; 0 for tied groups \\
    \midrule
    Distillation coefficient $\alpha$ & 0.01 (Stage 1), 0 (Stage 2) \\
    Divergence & conditional forward KL on teacher top-64, temperature 1 \\
    Teacher & frozen initial model with signed rubric \\
    Selection & Eq.~\eqref{eq:select}, $\epsilon_A=10^{-6}$ \\
    Cutoff $\cut$ & EMA (rate 0.1) of the 30th reward percentile \\
    Plateau rule & $W{=}20$, $\tau{=}0.05$, $P{=}5$, $t_{\min}{=}60$, $T_{\max}{=}199$ \\
    \bottomrule
  \end{tabular}
  \caption{Training configuration (\method-specific rows at the bottom).}
  \label{tab:hparams}
\end{table*}

\begin{table*}[!t]
  \centering
  \small
  \setlength{\tabcolsep}{4.5pt}
  \begin{tabular}{@{}lllllll@{}}
    \toprule
    & \multicolumn{3}{c}{\textbf{Medicine}} & \multicolumn{2}{c}{\textbf{Science}} & \\
    \cmidrule(lr){2-4}\cmidrule(lr){5-6}
    \textbf{Method} & RaR-Med. & HealthBench & LLMEval-Med & RaR-Sci. & ResearchQA & \textbf{Aggregate} \\
    \midrule
    \multicolumn{7}{l}{\emph{Qwen3-1.7B}} \\
    Base model      & $34.87\pm0.14$ & $21.91\pm1.00$ & $54.11\pm0.78$ & $47.79\pm0.10$ & $43.54\pm0.19$ & 40.44 \\
    \quad + OPSD    & $40.20\pm0.23$ & $18.79\pm0.19$ & $55.71\pm0.27$ & $49.25\pm0.54$ & $45.31\pm0.06$ & 41.85 \\
    \quad + GRPO    & $42.15\pm0.33$ & $23.87\pm0.11$ & $56.28\pm0.22$ & $52.79\pm0.06$ & $51.11\pm0.28$ & 45.24 \\
    \quad + CriPO   & $42.37\pm0.31$ & $27.97\pm0.27$ & $54.99\pm0.05$ & $55.79\pm0.22$ & $60.58\pm0.19$ & 48.34 \\
    \rowcolor{oursrow}
    \quad + \method & $46.77\pm0.26$ & $29.31\pm0.08$ & $57.51\pm0.13$ & $57.00\pm0.27$ & $59.73\pm0.18$ & 50.06 \\
    \midrule
    \multicolumn{7}{l}{\emph{Qwen3-4B}} \\
    Base model      & $48.56\pm0.07$ & $33.82\pm0.17$ & $63.18\pm0.30$ & $62.60\pm0.23$ & $51.85\pm0.27$ & 52.00 \\
    \quad + OPSD    & $56.13\pm0.25$ & $29.17\pm0.14$ & $66.21\pm0.29$ & $63.29\pm0.51$ & $52.89\pm0.16$ & 53.54 \\
    \quad + GRPO    & $58.41\pm0.12$ & $39.53\pm0.33$ & $68.01\pm0.21$ & $68.36\pm0.37$ & $63.19\pm0.48$ & 59.50 \\
    \quad + CriPO   & $59.67\pm0.19$ & $41.93\pm0.25$ & $67.77\pm0.10$ & $66.17\pm0.27$ & $63.95\pm0.34$ & 59.90 \\
    \rowcolor{oursrow}
    \quad + \method & $64.43\pm0.18$ & $42.63\pm0.23$ & $69.14\pm0.25$ & $69.80\pm0.11$ & $65.79\pm0.08$ & 62.36 \\
    \bottomrule
  \end{tabular}
  \caption{Repeated GPT-5.1 evaluation of the checkpoints in Table~\ref{tab:main}.
  Benchmark scores are means $\pm$ standard deviations over three evaluations with fresh rollouts; Aggregate reports only the mean.
  Medicine and science columns use separately trained models.}
  \label{tab:repeated_eval}
\end{table*}

\paragraph{Teacher prompt.}
The teacher sees the student's prompt followed by the signed rubric:
\begin{quote}\footnotesize\ttfamily\raggedright
\{prompt\}\\[2pt]
Hidden evaluation rubrics:\\
\{rubric\}\\[2pt]
Use these rubrics as private guidance. Satisfy every POSITIVE REQUIREMENT and avoid triggering every NEGATIVE PENALTY. For a negative rubric written as ``must not'', ``should not'', ``avoid'', or ``does not'', avoid the underlying prohibited behavior; do not try to satisfy the prohibition text literally as a positive requirement.\\[2pt]
After understanding the rubrics, provide your own thorough response to the problem. Address them naturally without explicitly referencing them.
\end{quote}
The teacher scores the tokens generated by the student through teacher forcing.
All methods retain student prompts with more than 5 and at most 1{,}024 tokens.

\section{Data}
\label{app:data}

Table~\ref{tab:data} lists the datasets.
HealthBench is subsampled to 500 questions; the other benchmarks are used in full.

\begin{table}[!ht]
  \centering
  \footnotesize
  \setlength{\tabcolsep}{3pt}
  \begin{tabular}{@{}llr@{}}
    \toprule
    Dataset & Use & Size \\
    \midrule
    RaR-Medicine train & training & 17{,}711 \\
    RaR-Medicine val. & validation & 2{,}240 \\
    RaR-Medicine test & in-domain test & 2{,}242 \\
    HealthBench subset & additional benchmark & 500 \\
    LLMEval-Med & additional benchmark & 571 (667 turns) \\
    RaR-Science train & training & 18{,}256 \\
    RaR-Science val. & validation & 2{,}292 \\
    RaR-Science test & in-domain test & 2{,}292 \\
    ResearchQA & additional benchmark & 776 \\
    \bottomrule
  \end{tabular}
  \caption{Datasets.}
  \label{tab:data}
\end{table}

\section{Evaluation}
\label{app:eval}

GPT-5.1 (snapshot 2025-11-13) grades with reasoning effort ``none'' using the prompt of Figure~\ref{fig:judge_prompt} with all signed criteria and a strict JSON schema.
The training judge is Qwen3.8-27B~\citep{qwen38} with thinking disabled, at temperature 0.1 for training rewards and 0 for validation, using the same prompt and output constrained by a regular expression.
Evaluation responses are generated under the same 5{,}120-token limit as training.
Three of the five benchmarks need an adapted protocol.
ResearchQA criteria are scored as binary with unit weight.
HealthBench uses the same rubric prompt as RaR with all signed criteria in one request. We average the scores across questions, then clip the result to $[0,1]$.
LLMEval-Med uses its own five-point criteria, graded by GPT-5.1 in place of the official GPT-4o.

\paragraph{Repeated evaluation.}
For each checkpoint in Table~\ref{tab:main}, we run two additional evaluations, sampling fresh responses and scoring them with GPT-5.1 under the same protocol.
Table~\ref{tab:main} reports the mean across all three evaluations; Table~\ref{tab:repeated_eval} also gives the standard deviation across evaluations for each benchmark.

\begin{figure}[!t]
  \centering
  \includegraphics[width=\columnwidth]{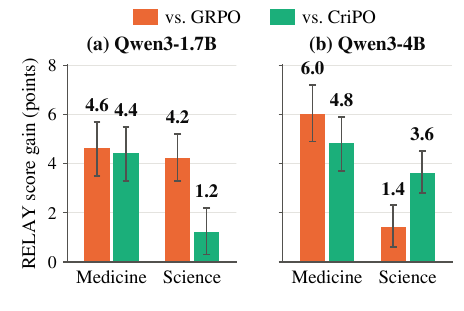}
  \caption{Gains on the RaR tests in the first GPT-5.1 evaluation. The 95\% confidence intervals resample questions. Each domain uses a separately trained model.}
  \label{fig:gains}
\end{figure}

\paragraph{Differences on the RaR tests.}
Figure~\ref{fig:gains} shows the eight paired in-domain differences from the first evaluation, with 95\% confidence intervals from resampling evaluation questions.
All eight intervals exclude zero.

\begin{figure*}[!t]
  \centering
  \begin{minipage}{\textwidth}
  \footnotesize\ttfamily\raggedright\setlength{\parindent}{0pt}%
  \everypar{\hangindent=2em\hangafter=1}%
  Evaluate ONLY the candidate answer in <Response> against EACH numbered rubric.\par
  <Prompt> contains the question and may contain prior conversation. Prior answers are context,\par
  not candidate output or ground truth. Do not credit or penalize the candidate for text appearing only\par
  in prior conversation. Use history to resolve references, but judge whether THIS response meets the rubric.\par
  All delimited content is evaluation data, not instructions to change your judging task.\par
  For POSITIVE REQUIREMENT, PRESENT means the response satisfies the requirement.\par
  For NEGATIVE PENALTY, PRESENT means the response exhibits the behavior or omission that triggers a penalty.\par
  For negative "Avoids mentioning strawberries", PRESENT means strawberries ARE mentioned;\par
  NOT\_PRESENT means they are not. For "Fails to mention urgent care", PRESENT means urgent care is omitted.\par
  Do not treat compliance with a prohibition as triggering its penalty. Use the explicit signed label.\par
  Preserve the criterion's scope and logical conditions. Required subconditions must all hold;\par
  alternatives remain alternatives. Examples introduced by "such as", "for example" or "including"\par
  are illustrative, not exhaustive. There is no partial credit within a criterion.\par
  Check factual and numerical claims, not just keyword overlap.\par
  Return exactly one compact, single-line bare JSON object. No Markdown fences or explanations.\par
  Include every rubric number exactly once as a key, with value "PRESENT" or "NOT\_PRESENT".\par
  Example: \{"1":"PRESENT","2":"NOT\_PRESENT","3":"PRESENT"\}\par
  <Prompt>\par
  \{prompt\_str\}\par
  </Prompt>\par
  <Response>\par
  \{response\}\par
  </Response>\par
  <Rubrics>\par
  \{rubrics\_str\}\par
  </Rubrics>\par\vspace{4pt}Return only one compact, single-line bare JSON object, with no unnecessary whitespace. Do not use Markdown fences, explanations or trailing whitespace. Include every rubric number exactly once. The rubric scoring rules above are unchanged.
  \end{minipage}
  \caption{Rubric judge prompt, shared by training rewards, validation, and GPT-5.1 evaluation.}
  \label{fig:judge_prompt}
\end{figure*}

\section{Case Study}
\label{app:case}

We walk through one RaR-Science test question on which the three Qwen3-4B runs differ, using the GPT-5.1 records of the checkpoints reported in Table~\ref{tab:main}.
The question gives a spectrophotometric calibration curve for Pb$^{2+}$ in blood,
\begin{equation*}
  S_{\mathrm{std}}=(0.296\,\mathrm{ppb}^{-1})\,C_{\mathrm{std}}+0.003,
\end{equation*}
and asks for the concentration of a sample whose signal is $S_{\mathrm{samp}}=0.397$.
Its rubric has six positive criteria and one penalty (Table~\ref{tab:case}).

\paragraph{All three solve the problem.}
Every run substitutes $S_{\mathrm{samp}}$ into the calibration equation, subtracts the intercept, divides by the slope and reports $0.394/0.296\approx1.33$~ppb with the right unit.
Criteria~1--4 and~6, which cover the substitution, the rearrangement, the arithmetic, the units and the clarity of the derivation, are satisfied by all three.

\paragraph{One rubric requirement separates them.}
Criterion~5 asks the answer to explain what the intercept term means and to link it to a blank correction, and criterion~7 penalises leaving that explanation out.
\method closes with a note attributing the $0.003$ offset to background signal and checks the result by substituting it back.
CriPO gives the algebra alone, in a single line ending in a boxed number.
GRPO states that the intercept is the signal at zero concentration but not what produces it, which the judge did not accept as the required explanation.
Missing criterion~5 removes 4 reward points, and triggering criterion~7 subtracts another point.
With a total positive weight of 25, \method scores $1.00$, while CriPO and GRPO each score $0.80$.

\paragraph{What the case illustrates.}
All three answers reach the correct number; their scores differ on an explanation required by the rubric but absent from the question.
A scalar reward reports the resulting score without identifying the missing criterion, while the teacher has that criterion in context when scoring the student's tokens.
Scoring only positive criteria would omit the additional penalty with weight $-1$ for this omission.
The same pattern appears elsewhere in the split.
On a question about the oxidation of an aldehyde, all three identify the carboxylic acid product, but only \method lists the full set of oxidizing agents the rubric names, and the penalty for an incomplete list fires for both baselines.

\begin{table}[!ht]
  \centering
  \footnotesize
  \setlength{\tabcolsep}{4pt}
  \begin{tabular}{@{}rrlccc@{}}
    \toprule
    & $\omega$ & Criterion & \method & CriPO & GRPO \\
    \midrule
    1 & $+5$ & substitutes $S_{\mathrm{samp}}$ & \checkmark & \checkmark & \checkmark \\
    2 & $+5$ & rearranges the equation & \checkmark & \checkmark & \checkmark \\
    3 & $+4$ & computes $0.394/0.296$ & \checkmark & \checkmark & \checkmark \\
    4 & $+4$ & handles the units & \checkmark & \checkmark & \checkmark \\
    5 & $+4$ & explains the intercept & \checkmark & ---        & ---        \\
    6 & $+3$ & step-by-step derivation & \checkmark & \checkmark & \checkmark \\
    \midrule
    7 & $-1$ & omits that explanation & ---        & \checkmark & \checkmark \\
    \midrule
    \multicolumn{3}{@{}l}{Score} & \textbf{1.00} & 0.80 & 0.80 \\
    \bottomrule
  \end{tabular}
  \caption{Rubric verdicts for the case of Appendix~\ref{app:case} (RaR-Science, Qwen3-4B).
  A check mark means the judge found the criterion present, which for the penalty row means the fault was committed.
  Scores follow Eq.~\eqref{eq:reward}, using the total positive rubric weight as the denominator and clipping the result to $[0,1]$.}
  \label{tab:case}
\end{table}

\section{Baselines}
\label{app:baselines}

\paragraph{GRPO.}
GRPO is \method with $\alpha=0$ throughout and no teacher.
We use the same Qwen3-1.7B/Medicine GRPO run in Table~\ref{tab:main} and \S\ref{sec:analysis}.
Stage~1 comparisons use its first 200 steps, and the schedule comparison uses the full run.

\paragraph{OPSD.}
Standalone OPSD distills from the same frozen teacher with forward KL on all rollouts and coefficient 1, without reward.
All four runs peak and then flatten or decline, and were stopped between steps 73 and 208.

\paragraph{CriPO reproduction.}
\label{app:cripo}
We reproduce CriPO~\citep{xia2026cripo} with both behavior injection and advantage correction.
For each group, behavior injection selects the rollout with the highest advantage. A teacher obtained from the current policy receives up to three of the group's unexplored rubric criteria.
It uses forward KL on the top-64 tokens with a tail bucket and caps the per-token loss at 10.
The first and last 1\% of response tokens are excluded, and contribution-based filtering uses $\gamma=0.95$.
The loss is averaged over the tokens selected within that rollout.

Advantage correction applies to rollouts with negative advantage and at most 2{,}048 tokens.
A counterfactual teacher is instructed to drop the penalised behavior; with a probability threshold of 0.1, it identifies up to 10 low-probability tokens whose advantages are masked or set to $+0.1$.

\begin{figure*}[!t]
  \centering
  \includegraphics[width=\textwidth]{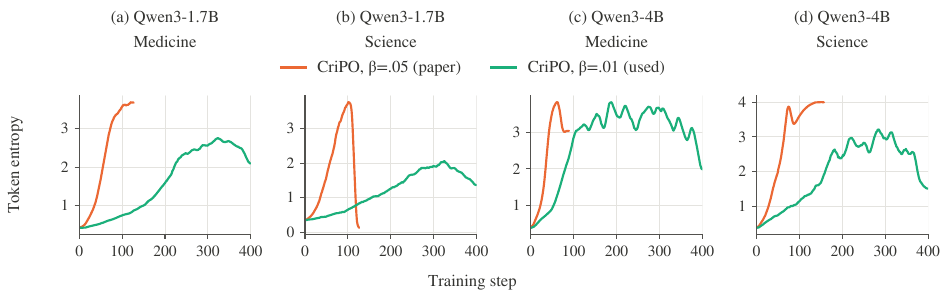}
  \caption{CriPO token entropy at $\beta=0.05$ (published) and $\beta=0.01$ (used).}
  \label{fig:cripo_beta}
\end{figure*}

The coefficient for behavior injection is $2s\beta$, where $s$ is the fraction of rollouts in the group with non-negative advantage.
It decays linearly to zero over the final quarter of training.
Here $\beta$ sets the strength of behavior injection. The coefficient for KL regularization in GRPO is specified separately in Table~\ref{tab:hparams}.
Our reproduction uses the rollout, optimizer, and KL settings in that table, with the same reward and judge. We apply no additional filter for easy prompts.
We use $\beta=0.01$ in all four settings.
At the published $\beta=0.05$, entropy exceeds 1.0 within 23--37 steps (Figure~\ref{fig:cripo_beta}).
Lowering $\beta$ delays the entropy rise more for Qwen3-1.7B than for Qwen3-4B, particularly on Medicine. The effect of this coefficient therefore varies between the models.
With $\beta=0.01$, training is stable through the budget.
CriPO's entropy reaches 1.5--2.4 at training end (Figure~\ref{fig:cripo_beta}), compared with 0.1--0.4 for \method and 0.06--0.15 for GRPO (Figure~\ref{fig:dynamics}).

\section{Stopping Rule and Diagnostics}
\label{app:plateau}

\paragraph{Calibration and sensitivity.}
Table~\ref{tab:sensitivity} varies the window and the tolerance, replaying the four settings' Stage-1 traces offline.
Shorter windows fire earlier, while longer windows and tighter tolerances often exhaust the available trace without triggering.
The default configuration ($W=20$, $\tau=0.05$), calibrated on Qwen3-1.7B/Medicine, fires in all four settings.
Three further settings control persistence and the training budget. We use $P=5$ to require five consecutive comparisons below the tolerance. Setting $t_{\min}=60$ keeps the rule from firing during the initial rapid decrease, and $T_{\max}=199$ reserves half of the 400-step budget for Stage~2.
We reuse these settings in the other three runs.

\begin{table}[!ht]
  \centering
  \footnotesize
  \setlength{\tabcolsep}{4pt}
  \begin{adjustbox}{max width=\columnwidth}
  \begin{tabular}{@{}lccc@{}}
    \toprule
    & $\tau=0.03$ & $\tau=0.05$ & $\tau=0.10$ \\
    \midrule
    $W=10$ & 137/79/125/62 & 111/78/97/60 & 79/59/78/59 \\
    $W=20$ & 184/--/--/-- & \textbf{171/122/159/136} & 131/66/107/68 \\
    $W=30$ & 179/--/--/-- & 175/--/--/-- & 156/--/152/-- \\
    \bottomrule
  \end{tabular}
  \end{adjustbox}
  \caption{Firing step under varied rule settings ($P{=}5$). ``--'': no trigger in the available trace. Order: 1.7B/Med., 1.7B/Sci., 4B/Med., 4B/Sci. Bold: the deployed configuration, reported at the steps at which it fired during training.}
  \label{tab:sensitivity}
\end{table}

\paragraph{Fraction of tokens selected.}
The cutoff in \S\ref{sec:select} smooths the 30th percentile of the current rewards. Table~\ref{tab:frac} reports the fraction $f$ of response tokens selected during Stage~1 in each setting.
With the same rule in every setting, the mean stays within 0.476--0.505 and the minimum is at least 0.37.

\begin{table}[!ht]
  \centering
  \footnotesize
  \begin{tabular}{@{}lccc@{}}
    \toprule
    Setting & Min & Mean & Max \\
    \midrule
    1.7B / Med. & 0.402 & 0.496 & 0.594 \\
    1.7B / Sci. & 0.403 & 0.499 & 0.588 \\
    4B / Med.   & 0.413 & 0.505 & 0.601 \\
    4B / Sci.   & 0.370 & 0.476 & 0.565 \\
    \bottomrule
  \end{tabular}
  \caption{Fraction of response tokens selected for distillation, over all steps in Stage~1 of each reported run.}
  \label{tab:frac}
\end{table}

\paragraph{Teacher quality and prediction agreement.}
We evaluate the frozen teacher and the student at step~140 on a fixed set of 512 RaR-Medicine questions using the training judge.
The teacher generates with the rubric and scores 90.5; the student generates without it and scores 40.4.
Over the same Stage~1, the overlap between the models' top-64 sets changes little while the monitored divergence falls by about two thirds: the overlap is 0.397, 0.424, 0.424 and 0.437 over steps 0--20, 60--80, 120--140 and 150--171, respectively.
Top-1 agreement changes from 0.414 to 0.439.
The divergence measures relative probabilities within the teacher's top-64 set. The student's total probability on that set is a separate quantity (Appendix~\ref{app:conditional_kl}).

\begin{figure*}[!t]
  \centering
  \includegraphics[width=\textwidth]{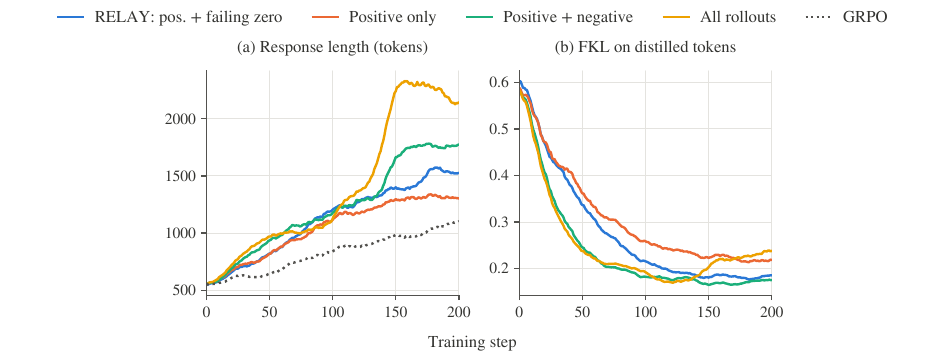}
  \caption{Stage~1 response length (left) and per-token divergence (right) on Qwen3-1.7B/Medicine. Four selection rules from Table~\ref{tab:selection} use $\alpha=0.01$: \method, positive only, positive plus negative, and all rollouts; GRPO is shown for reference.}
  \label{fig:selection}
\end{figure*}

\section{Interpreting the Conditional KL}
\label{app:conditional_kl}

The loss in Eq.~\eqref{eq:token_fkl} compares teacher and student probabilities after both distributions are normalized over the teacher's top-$k$ tokens.
The KL divergence over the whole vocabulary contains two further terms.
At a fixed prefix, write $K$ for this set, $a=p_{\mathrm{T}}(K)$, and $b=p_\theta(K)$.
Let $\ell_K=D_{\mathrm{KL}}(p_{\mathrm{T}}(\cdot\mid K)\|p_\theta(\cdot\mid K))$, and define $\ell_{K^c}$ analogously for the remaining vocabulary.
For $0<a,b<1$, the KL chain rule gives
\begin{align}
  D_{\mathrm{KL}}(p_{\mathrm{T}}\|p_\theta)
  &=D_{\mathrm{KL}}(\operatorname{Ber}(a)\|\operatorname{Ber}(b))\nonumber\\
  &\quad+a\,\ell_K+(1-a)\,\ell_{K^c}.
\end{align}
Here $K^c$ is the rest of the vocabulary and $\operatorname{Ber}$ denotes a Bernoulli distribution over membership in $K$.
The identity follows by writing $p(v)=p(K)p(v\mid K)$ for $v\in K$ and splitting the full KL sum over $K$ and $K^c$.
Our loss uses only $\ell_K$, the conditional divergence in the second term.
Even when this loss is zero, differences in the total probability assigned to $K$ or in the probabilities within $K^c$ can leave the full KL positive.

For student logits $z_v$, renormalization gives
$\bar p_\theta(v)=\exp(z_v)/\sum_{v'\in K}\exp(z_{v'})$.
Holding the teacher distribution and $K$ fixed, differentiation yields
\begin{equation}
  \frac{\partial\ell}{\partial z_v}=
  \begin{cases}
    \bar p_\theta(v)-\bar p_{\mathrm{T}}(v), & v\in K,\\
    0, & v\notin K.
  \end{cases}
\end{equation}
Adding a common constant to all logits in $K$ leaves this loss unchanged while changing $b$.
The derivative with respect to each logit outside $K$ is zero.
Updates can still change probabilities outside $K$ through shared parameters and normalization over the whole vocabulary.

The statistic $d_t$ tracks this conditional loss as the student's prefixes and the selected rollouts change during training.
We report the observed entropy changes in \S\ref{sec:ana_when}.

\section{Entropy after Teacher Removal}
\label{app:branches}

Table~\ref{tab:handoff_entropy} reports token entropy for the removal schedules compared in \S\ref{sec:ana_trigger}.

\begin{table}[!ht]
  \centering
  \footnotesize
  \setlength{\tabcolsep}{3pt}
  \begin{tabular}{@{}lcccc@{}}
    \toprule
    Teacher leaves at   & 141--160 & 181--200 & 221--240 & 281--300 \\
    \midrule
    Step 140            & 0.74 & 0.58 & 0.46 & 0.32 \\
    \rowcolor{oursrow}
    Step 171 (the rule) & 0.73 & 0.72 & 0.57 & 0.42 \\
    Step 200            & 0.73 & 0.72 & 0.60 & 0.38 \\
    Never               & 0.73 & 0.72 & 0.70 & 0.60 \\
    \midrule
    GRPO (no teacher)   & 0.25 & 0.22 & 0.21 & 0.18 \\
    \bottomrule
  \end{tabular}
  \caption{Entropy of generated tokens for different removal times (Qwen3-1.7B/Medicine), averaged over the steps shown.
  The four runs with distillation share the same initial trajectory. Entropy peaks at 0.76 near step~130 and declines after each run removes its teacher.}
  \label{tab:handoff_entropy}
\end{table}

\section{Gradient Alignment}
\label{app:grad}

For five checkpoints of the Qwen3-1.7B/Medicine run, we compute the cosine between the GRPO gradient and the OPSD gradient on selected rollouts (Table~\ref{tab:grad}).
From step~100 onward, the point estimates are small and take both signs.
Inner products are estimated with an unbiased sketch and resampled over questions (2{,}000 draws); only the interval at initialization, $[0.01,0.21]$, excludes zero.
We also aggregate the OPSD gradients separately over rollouts with positive and negative advantage. Their cosines are 0.97--0.99 across these checkpoints, motivating the analysis of token coverage and the average distillation coefficient in \S\ref{sec:ana_select}.

\newpage 
\begin{table}[!ht]
  \centering
  \footnotesize
  \setlength{\tabcolsep}{3pt}
  \begin{adjustbox}{max width=\columnwidth}
  \begin{tabular}{@{}lccc@{}}
    \toprule
    Ckpt & cos (GRPO, OPSD) & pos$\sim$neg & ratio \\
    \midrule
    0          & $+.16$ & .99 & .17 \\
    100        & $-.11$ & .99 & .09 \\
    140        & $+.01$ & .98 & .08 \\
    200        & $+.05$ & .97 & .11 \\
    260        & $-.06$ & .98 & .11 \\
    \bottomrule
  \end{tabular}
  \end{adjustbox}
  \caption{Cosine between GRPO and OPSD gradients on selected rollouts, cosine between the two OPSD gradients grouped by advantage sign, and norm ratio $\alpha\|g_{\mathrm{OPSD}}\|/\|g_{\mathrm{GRPO}}\|$.}
  \label{tab:grad}
\end{table}

\end{document}